\documentclass[11pt,a4paper]{article}
\usepackage[hyperref]{acl2019}
\usepackage{times}
\usepackage{latexsym}
\usepackage{graphicx}
\usepackage[colorinlistoftodos]{todonotes}
\usepackage{amsfonts}
\usepackage{amsmath}
\usepackage{svg}
\usepackage{url}
\usepackage{booktabs}

\aclfinalcopy % Uncomment this line for the final submission
\title{Fine-tuning Pre-Trained Transformer Language Models to Distantly Supervised Relation Extraction}

\author{Christoph Alt ~~~~~ Marc H\"ubner ~~~~~
Leonhard Hennig \\
\mbox{}\\
German Research Center for Artificial Intelligence (DFKI)\\
Speech and Language Technology Lab \\
\texttt{\{christoph.alt, marc.huebner, leonhard.hennig\}@dfki.de}}

\date{}
\begin{document}
\maketitle

\begin{abstract}
Distantly supervised relation extraction is widely used to extract relational facts from text, but suffers from noisy labels. Current relation extraction methods try to alleviate the noise by multi-instance learning and by providing supporting linguistic and contextual information to more efficiently guide the relation classification. While achieving state-of-the-art results, we observed these models to be biased towards recognizing a limited set of relations with high precision, while ignoring those in the long tail. To address this gap, we utilize a pre-trained language model, the OpenAI Generative Pre-trained Transformer (GPT)~\cite{radford_improvinglu_2018}. The GPT and similar models have been shown to capture semantic and syntactic features, and also a notable amount of ``common-sense'' knowledge, which we hypothesize are important features for recognizing a more diverse set of relations. By extending the GPT to the distantly supervised setting, and fine-tuning it on the NYT10 dataset, we show that it predicts a larger set of distinct relation types with high
confidence. Manual and automated evaluation of our model shows that it achieves a state-of-the-art AUC score of 0.422 on the NYT10 dataset, and performs especially well at higher recall levels.

\end{abstract}

\section{Introduction}
\label{Introduction}
\begin{figure}[t!]
\centering
\includegraphics[width=\linewidth]{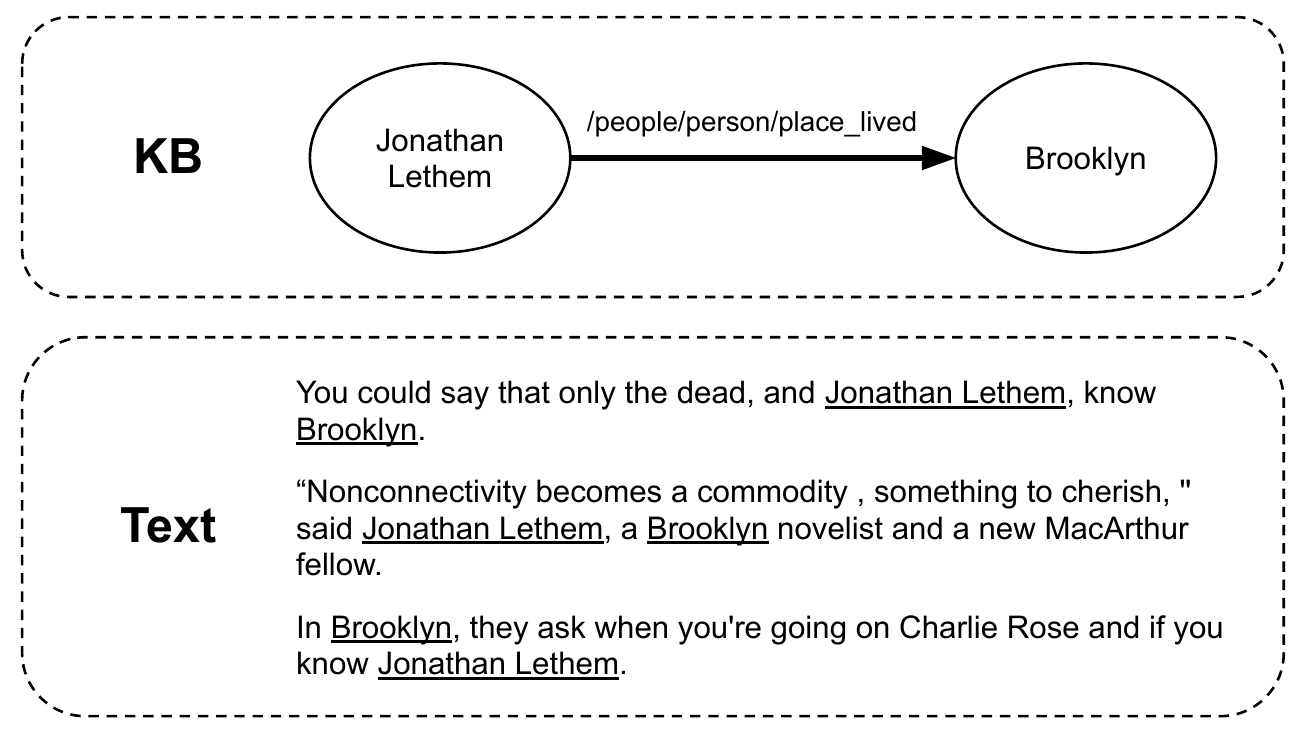}
\caption{Distant supervision generates noisily labeled relation mentions by aligning entity tuples in a text corpus with relation instances from a knowledge base.}
\label{fig:re_example}
\end{figure}

Relation extraction (RE), defined as the task of identifying the relationship between concepts mentioned in text, is a key component of many natural language processing applications, such as knowledge base population~\cite{tac_kbp_2010} and question answering~\cite{yu_improvednr_2017}. 
Distant supervision~\cite{mintz_distantsf_2009,hoffmann_knowledge_based_2011} is a popular approach to heuristically generate labeled data for training RE systems by aligning entity tuples in text with known relation instances from a knowledge base, but suffers from noisy labels and incomplete knowledge base information~\cite{min_distant_2013,fan_distant_2014}. Figure \ref{fig:re_example} shows an example of three sentences labeled with an existing KB relation, two of which are false positives and do not actually express the relation. 

Current state-of-the-art RE methods try to address these challenges by applying multi-instance learning  methods~\cite{mintz_distantsf_2009,surdeanu_multi_instance_2012,lin_neuralre_2016} and guiding the model by \textit{explicitly} provided semantic and syntactic knowledge, e.g.\ part-of-speech tags~\cite{zeng_relationcv_2014} and dependency parse information~\cite{surdeanu_multi_instance_2012,zhang_graphco_2018}.
Recent methods also utilize side information, e.g.\ paraphrases, relation aliases, and entity types~\cite{reside_re}. However, we observe that these models are often biased towards recognizing a limited set of relations with high precision, while ignoring those in the long tail (see Section~\ref{subsec:manual_eval}).

Deep language representations, e.g.\ those learned by the Transformer~\cite{vaswani_attention_2017} via language modeling \cite{radford_improvinglu_2018}, have been shown to \textit{implicitly} capture useful semantic and syntactic properties of text solely by unsupervised pre-training~\cite{peters_deepcw_2018}, as demonstrated by state-of-the-art performance on a wide range of natural language processing tasks~\cite{vaswani_attention_2017,peters_deepcw_2018,radford_improvinglu_2018,devlin_bert_2018}, including supervised relation extraction~\cite{alt_improving_2019}. \citet{radford_gpt2} even found language models to perform fairly well on answering open-domain questions without being trained on the actual task, suggesting they capture a limited amount of ``common-sense'' knowledge. We hypothesize that pre-trained language models provide a stronger signal for distant supervision, better guiding relation extraction based on the knowledge acquired during unsupervised pre-training. Replacing explicit linguistic and side-information with implicit features improves domain and language independence and could increase the diversity of the recognized relations.

In this paper, we introduce a Distantly Supervised Transformer for Relation Extraction (DISTRE).
We extend the standard Transformer architecture by a selective attention mechanism to handle multi-instance learning and prediction, which allows us to fine-tune the pre-trained Transformer language model 
directly on the distantly supervised RE task. This minimizes explicit feature extraction and reduces the risk of error accumulation. In addition, the self-attentive architecture allows the model to efficiently capture long-range dependencies and the language model to utilize knowledge about the relation between entities and concepts acquired during unsupervised pre-training. Our model achieves a state-of-the-art AUC score of $0.422$ on the NYT10 dataset, and performs especially well at higher recall levels, when compared to competitive baseline models.

We selected the GPT as our language model because of its fine-tuning efficiency and reasonable hardware requirements, compared to e.g.\  LSTM-based language models \cite{ruder_universallm_2018, peters_deepcw_2018} or BERT \cite{devlin_bert_2018}. The contributions of this paper can be summarized as follows:
\begin{itemize}
    \item We extend the GPT to handle bag-level, multi-instance training and prediction for distantly supervised datasets, by aggregating sentence-level information with selective attention to produce bag-level predictions (\S~\ref{sec:mi_learning_transformer}).
    \item We evaluate our fine-tuned language model on the NYT10 dataset and show that it achieves a state-of-the-art AUC compared to RESIDE~\cite{reside_re} and PCNN+ATT~\cite{lin_neuralre_2016} in held-out evaluation (\S~\ref{sec:experiments},\ \S~\ref{subsec:held_out_eval}).
    \item We follow up on these results with a manual evaluation of ranked predictions, demonstrating that our model predicts a more diverse set of relations and performs especially well at higher recall levels (\S~\ref{subsec:manual_eval}).
    \item We make our code publicly available at \url{https://github.com/DFKI-NLP/DISTRE}.
\end{itemize}

\section{Transformer Language Model}
\label{sec:distre}
This section reviews the Transformer language model as introduced by \citet{radford_improvinglu_2018}. We first define the Transformer-Decoder (Section~\ref{sec:model_architecture}), followed by an introduction on how contextualized representations are learned with a language modeling objective (Section~\ref{sec:unsuperv_pretraining}).

\subsection{Transformer-Decoder}
\label{sec:model_architecture}
\begin{figure}[t]
\centering
% architecture drawing: https://docs.google.com/drawings/d/16--THTwBC7I-xFUo4wFYmmaAKOidBgfmQ0382z1j7wA/edit
\includegraphics[width=0.8\textwidth, trim={0cm 8cm 2cm 0cm},clip]{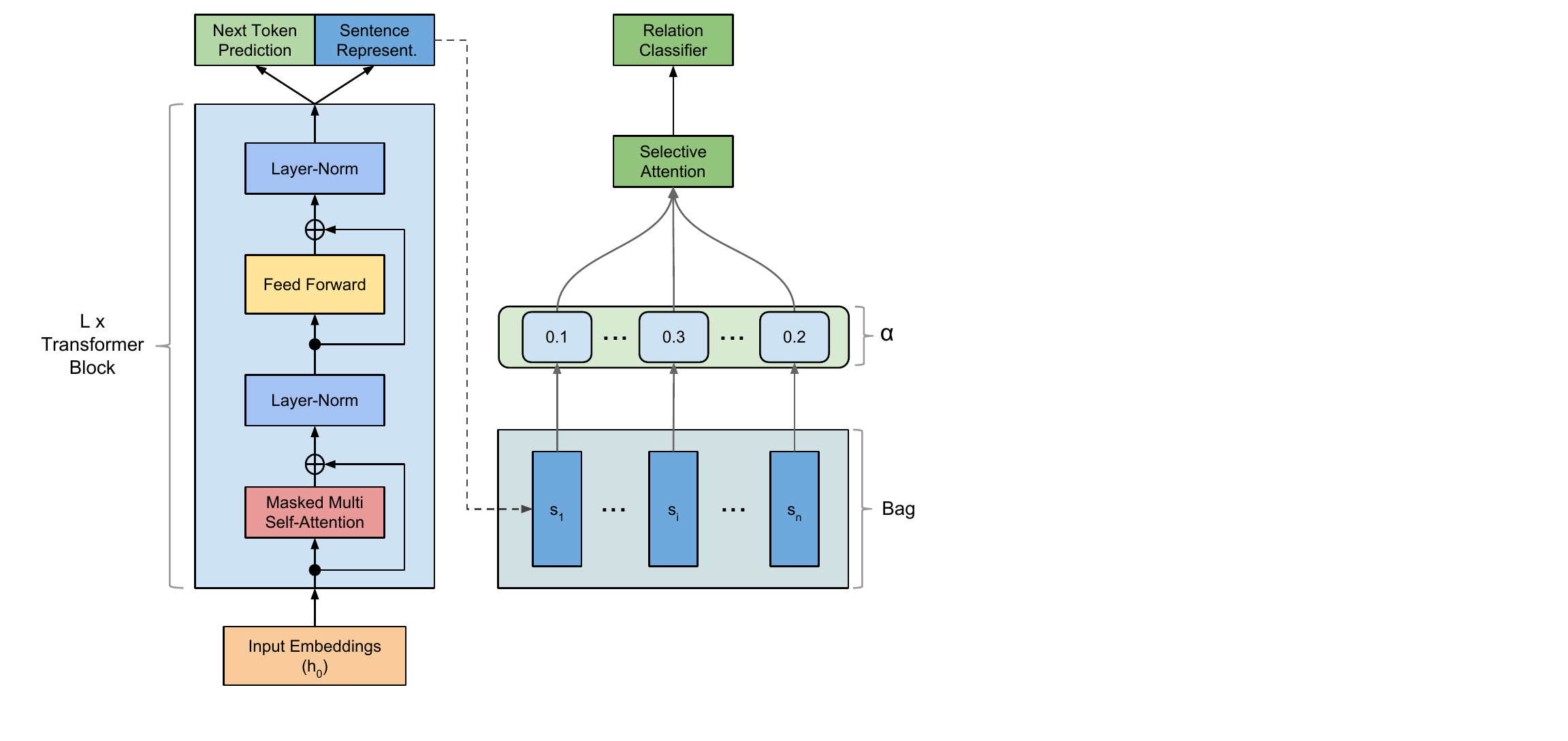}
\caption{Transformer-Block architecture and training objectives. A Transformer-Block is applied at each of the $L$ layers to produce states $h_1$ to $h_L$. After encoding each sentence in a bag into its representation $s_i$, selective attention informs the relation classifier with a representation aggregated over all sentences $[s_1, \ldots, s_n]$.}
\label{fig:model_architecture}
\end{figure}

The Transformer-Decoder~\cite{liu_generatingwb_2018}, shown in Figure~\ref{fig:model_architecture}, is a decoder-only variant of the original Transformer~\cite{vaswani_attention_2017}. Like the original Transformer, the model repeatedly encodes the given input representations over multiple layers (i.e., Transformer blocks), consisting of masked multi-head self-attention followed by a position-wise feedforward operation. In contrast to the original decoder blocks this version contains no form of unmasked self-attention since there are no encoder blocks. This is formalized as follows:
\begin{equation}
    \begin{tabular}{l}
        $h_0 = T W_e + W_p$ \\
        $h_l = tf\_block(h_{l-1})\ \forall\ l\ \in\ [1, L]$ \\
    \end{tabular}
\end{equation}
Where $T$ is a matrix of one-hot row vectors of the token indices in the sentence, $W_e$ is the token embedding matrix, $W_p$ is the positional embedding matrix, $L$ is the number of Transformer blocks, and $h_l$ is the state at layer $l$. Since the Transformer has no implicit notion of token positions, the first layer adds a learned positional embedding $e_p \in \mathbb{R}^d$ to each token embedding $e^p_t \in \mathbb{R}^d$ at position $p$ in the input sequence. The self-attentive architecture allows an output state $h^p_l$ of a block to be informed by all input states $h_{l-1}$, which is key to efficiently model long-range dependencies. For language modeling, however, self-attention must be constrained (masked) not to attend to positions ahead of the current token. 
For a more exhaustive description of the architecture, we refer readers to~\citet{vaswani_attention_2017} and the excellent guide ``The Annotated Transformer''.\footnote{\url{http://nlp.seas.harvard.edu/2018/04/03/attention.html}}

\subsection{Unsupervised Pre-training of Language Representations}
\label{sec:unsuperv_pretraining}

Given a corpus $\mathcal{C} = \{c_1, \ldots, c_n\}$ of tokens $c_i$, the language modeling objective maximizes the likelihood
\begin{equation}
    \label{eq:lm_objective}
    L_1(\mathcal{C}) = \sum_i \log P(c_i | c_{i-1}, \ldots, c_{i-k}; \theta),
\end{equation}
where k is the context window considered for predicting the next token $c_i$ via the conditional probability $P$. The distribution over the target tokens is modeled using the previously defined Transformer model as follows:
\begin{equation}
    P(c) = softmax(h_L W^T_e),
\end{equation}
where $h_L$ is the sequence of states after the final layer $L$, $W_e$ is the embedding matrix, and $\theta$ are the model parameters that are optimized by stochastic gradient descent. This results in a probability distribution for each token in the input sequence.

\section{Multi-Instance Learning with the Transformer}
\label{sec:mi_learning_transformer}
This section introduces our extension to the original transformer architecture, enabling bag-level multi-instance learning on distantly supervised datasets (Section~\ref{sec:dist_superv_re}), followed by a description of our task-specific input representation for relation extraction (Section~\ref{sec:input_representation}).

\subsection{Distantly Supervised Fine-tuning on Relation Extraction}
\label{sec:dist_superv_re}
After pre-training with the objective in Eq.~\ref{eq:lm_objective}, the language model is fine-tuned on the relation extraction task. We assume a labeled dataset $\mathcal{D} = \{(x_i, head_i, tail_i, r_i)\}^{N}_{i=1}$, where each example consists of an input sequence of tokens $x_i = [x^1, \ldots, x^m]$, the positions $head_i$ and $tail_i$ of the relation's head and tail entity in the sequence of tokens, and the corresponding relation label $r_i$, assigned by distant supervision. Due to its noisy annotation, label $r_i$ is an unreliable target for training. Instead, the relation classification is applied on a bag level, representing each entity pair $(head, tail)$ as a set $S = \{x_1, \ldots, x_n\}$ consisting of all sentences that contain the entity pair. A set representation $s$ is then derived as a weighted sum over the individual sentence representations:
\begin{equation}
    s = \sum_i \alpha_i s_i,
\end{equation}
where $\alpha_i$ is the weight assigned to the corresponding sentence representation $s_i$. A sentence representation is obtained by feeding the token sequence $x_i$ of a sentence to the pre-trained model and using the last state $h^m_L$ of the final state representation $h_L$ as its representation $s_i$. The set representation $s$ is then used to inform the relation classifier.

We use selective attention~\cite{lin_neuralre_2016}, shown in Figure~\ref{fig:model_architecture}, as our approach for aggregating a bag-level representation $s$ based on the individual sentence representations $s_i$. Compared to average selection, where each sentence representation contributes equally to the bag-level representation, selective attention learns to identify the sentences with features most clearly expressing a relation, while de-emphasizing those that contain noise. The weight $\alpha_i$ is obtained for each sentence by comparing its representation against a learned relation representation $r$:

\begin{equation}
    \alpha_i = \frac{exp(s_i r)}{\sum_{j=1}^n exp(s_j r)}
\end{equation}

To compute the output distribution $P(l)$ over relation labels, a linear layer followed by a softmax is applied to $s$:
\begin{equation}
    P(l | S, \theta) = softmax(W_r s + b),
\end{equation}
where $W_r$ is the representation matrix of relations $r$ and $b \in R^{d_r}$ is a bias vector.
During fine-tuning we want to optimize the following objective:
\begin{equation}
    \label{eq:finetune_objective}
    L_2(\mathcal{D}) = \sum^{|\mathcal{S}|}_{i=1} \log P(l_i | S_i, \theta)
\end{equation}
According to \citet{radford_improvinglu_2018}, introducing language modeling as an auxiliary objective during fine-tuning improves generalization and leads to faster convergence. Therefore, our final objective combines Eq.~\ref{eq:lm_objective} and Eq.~\ref{eq:finetune_objective}:
\begin{equation}
    L(\mathcal{D}) = \lambda * L_1(\mathcal{D}) + L_2(\mathcal{D}),
\end{equation}
where the scalar value $\lambda$ is the weight of the language model objective during fine-tuning.

\subsection{Input Representation}
\label{sec:input_representation}
\begin{figure}[t!]
\centering
\includegraphics[width=0.8\textwidth, trim={0cm 9cm 3cm 0cm},clip]{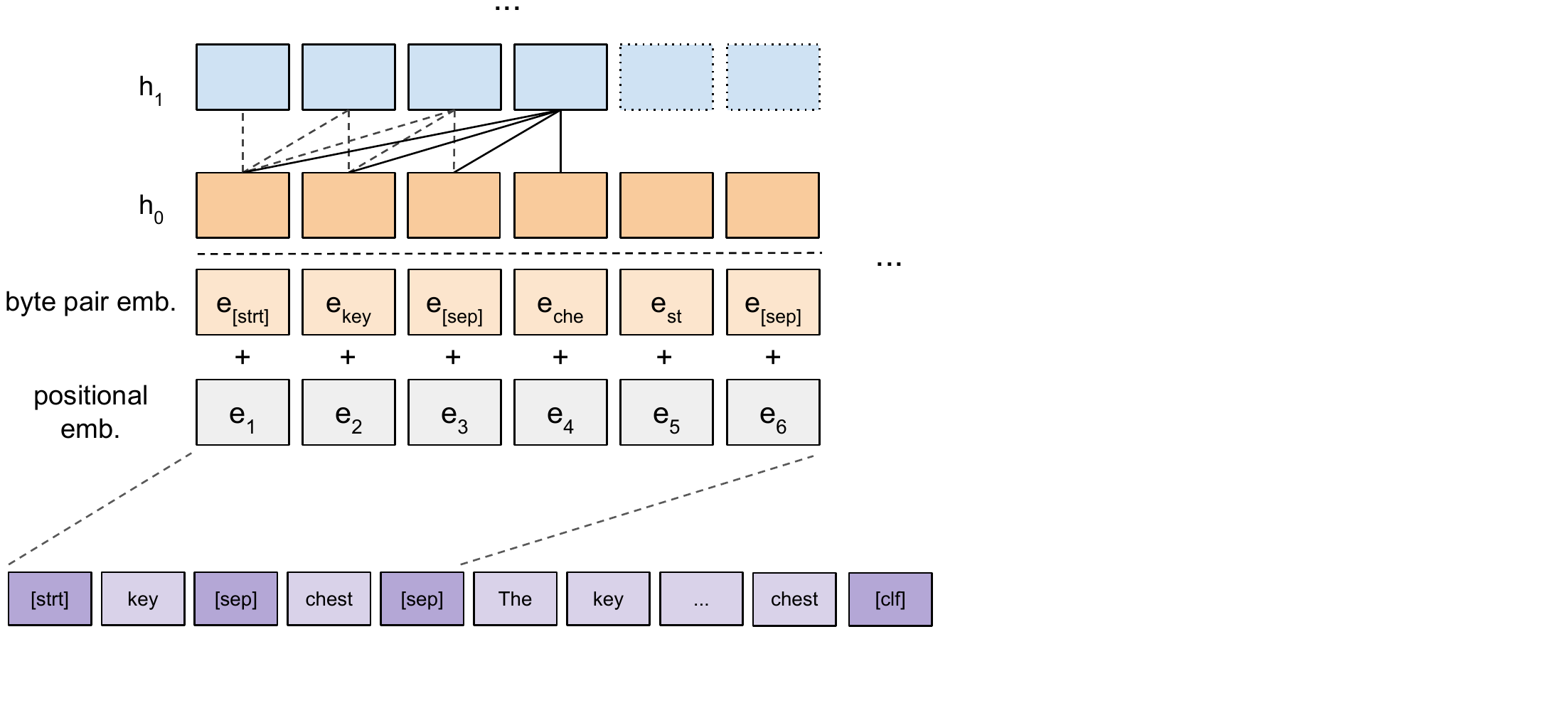}
\caption{Relation extraction requires a structured input for fine-tuning, with special delimiters to assign different meanings to parts of the input. The input embedding $h_0$ is created by summing over the positional embedding and the byte pair embedding for each token. States $h_l$ are obtained by self-attending over the states of the previous layer $h_{l-1}$.}
\label{fig:distre_input_format}
\end{figure}

Our input representation (see Figure~\ref{fig:distre_input_format}) encodes each sentence as a sequence of tokens. To make use of sub-word information, we tokenize the input text using byte pair encoding (BPE)~\cite{sennrich_subword_2016}. The BPE algorithm creates a vocabulary of sub-word tokens, starting with single characters. Then, the algorithm iteratively merges the most frequently co-occurring tokens into a new token until a predefined vocabulary size is reached. For each token, we obtain its input representation by summing over the corresponding token embedding and positional embedding.

While the model is pre-trained on plain text sentences, relation extraction requires a structured input, namely a sentence and relation arguments.
To avoid task-specific changes to the architecture, we adopt a traversal-style approach similar to~\citet{radford_improvinglu_2018}. The structured, task-specific input is converted to an ordered sequence to be directly fed to the model without architectural changes. Figure~\ref{fig:distre_input_format} provides a visual illustration of the input format. It starts with the tokens of the head and tail entity, separated by delimiters, followed by the token sequence of the sentence containing the entity pair, and ends with a special classification token. The classification token signals the model to generate a sentence representation for relation classification. Since our model processes the input left-to-right, we add the relation arguments to the beginning, to bias the attention mechanism towards their token representation while processing the sentence's token sequence.

\section{Experiment Setup}
\label{experiment_setup}
In the following section we describe our experimental setup. We run our experiments on the distantly supervised NYT10 dataset and use PCNN+ATTN~\cite{lin_neuralre_2016} and RESIDE~\cite{reside_re} as the state-of-the-art baselines.

The piecewise convolutional neural network (PCNN) segments each input sentence into parts to the left, middle, and right of the entity pair, followed by convolutional encoding and selective attention to inform the relation classifier with a bag-level representation. RESIDE, on the other hand, uses a bidirectional gated recurrent unit (GRU) to encode the input sentence, followed by a graph convolutional neural network (GCN) to encode the explicitly provided dependency parse tree information. This is then combined with named entity type information to obtain a sentence representation that can be aggregated via selective attention and forwarded to the relation classifier. 

\label{sec:experiments}
\subsection{NYT10 Dataset}
\label{subsec:exp_datasets}
The NYT10 dataset by~\citet{riedel_modeling_2010} is a standard benchmark for distantly supervised relation extraction. It was generated by aligning Freebase relations with the New York Times corpus, with the years 2005--2006 reserved for training and 2007 for testing. We use the version of the dataset pre-processed by~\citet{lin_neuralre_2016}, which is openly accessible online.\footnote{\url{https://drive.google.com/file/d/1eSGYObt-SRLccvYCsWaHx1ldurp9eDN_}} The training data contains 522,611 sentences, 281,270 entity pairs and 18,252 relational  facts. The test data contains 172,448 sentences, 96,678 entity pairs and 1,950 relational facts. There are 53 relation types, including NA if no relation holds for a given sentence and entity pair. Per convention we report \emph{Precision@N} (precision scores for the top 100, top 200, and top 300 extracted relation instances) and a plot of the precision-recall curves. Since the test data is also generated via distant supervision, and can only provide an approximate measure of the performance, we also report \emph{P@100}, \emph{P@200}, and \emph{P@300} based on a manual evaluation.

\subsection{Pre-training}
\label{subsec:exp_pretraining}
Since pre-training is computationally expensive, and our main goal is to show its effectiveness by fine-tuning on the distantly supervised relation extraction task, we reuse the language model\footnote{\url{https://github.com/openai/finetune-transformer-lm}} published by~\citet{radford_improvinglu_2018} for our experiments. The model was trained on the BooksCorpus~\cite{zhu_aligningba_2015}, which contains around 7,000 unpublished books with a total of more than 800M words of different genres. The model consists of $L=12$ decoder blocks with 12 attention heads and 768 dimensional states, and a feed-forward layer of 3072 dimensional states. We reuse the byte-pair encoding vocabulary of this model, but extend it with task-specific tokens (e.g., start, end, delimiter).

\subsection{Hyperparameters}
\label{sec:exp_hyperparams}
During our experiments we found the hyperparameters for fine-tuning, reported in~\citet{radford_improvinglu_2018}, to be very effective. We used the Adam optimization scheme~\cite{adam_2015} with $\beta_1=0.9$, $\beta_2=0.999$, a batch size of 8, a learning rate of 6.25e-5, and a linear learning rate decay schedule with warm-up over 0.2\% of training updates. We trained the model for 3 epochs and applied residual and attention dropout with a rate of 0.1, and classifier dropout with a rate of 0.2.

\section{Results}
\label{sec:results}
This section presents our experimental results. We compare DISTRE to other works on the NYT10 dataset, and show that it recognizes a more diverse set of relations, while still achieving state-of-the-art AUC. Even without explicitly provided side information and linguistic features.

\subsection{Held-out Evaluation}
\label{subsec:held_out_eval}

\begin{figure}[t!]
\centering
\includegraphics[width=\linewidth]{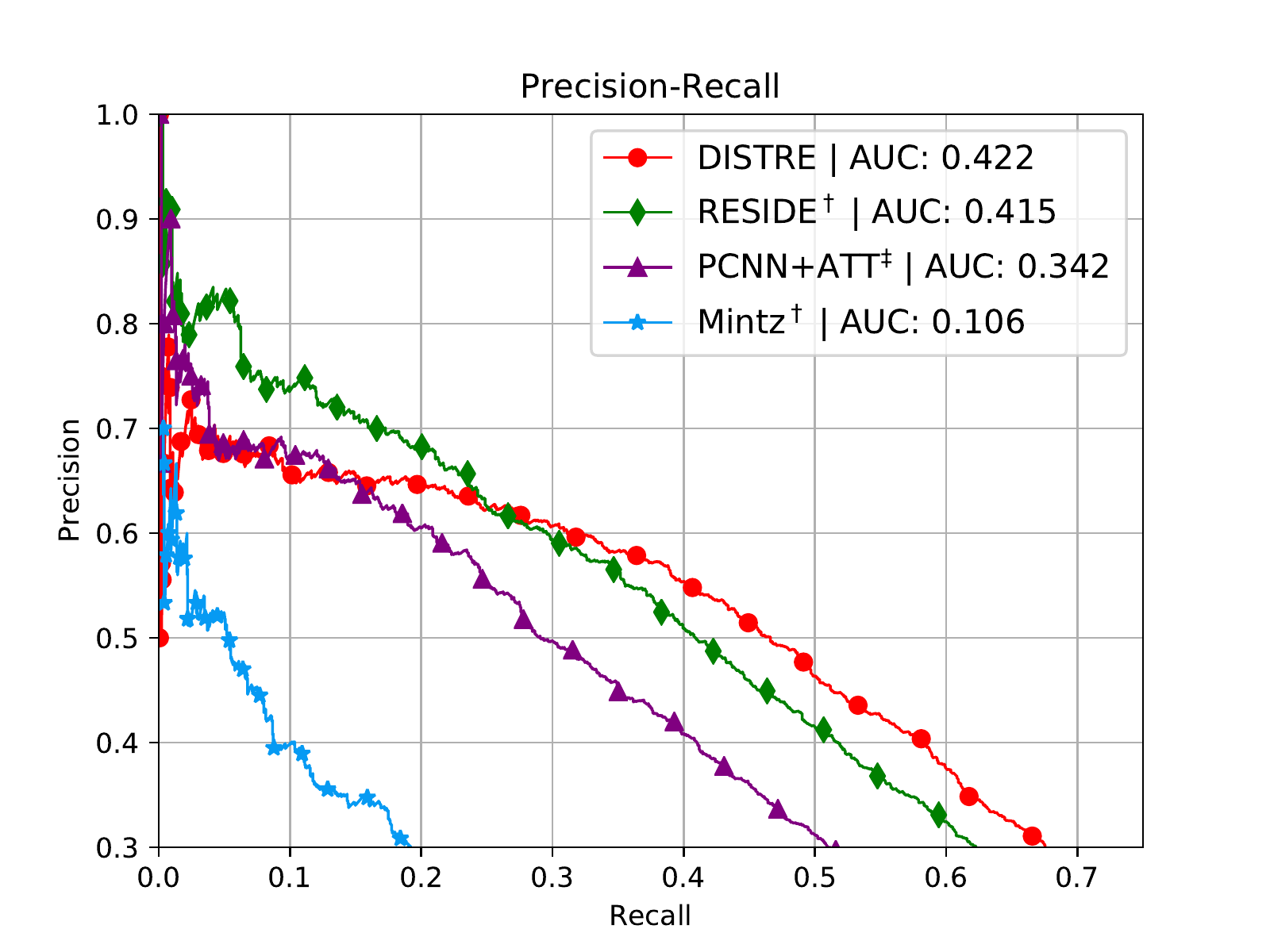}
\caption{Precision-Recall curve on the NYT dataset. Our method (DISTRE) shows a more balanced performance across relations, especially in the long tail. $\mathrm{\dagger}$ marks results reported by \citet{reside_re}. $\mathrm{\ddagger}$ indicates results we obtained with the OpenNRE\footnotemark\ implementation.}
\label{fig:nyt_pr_curve}
\end{figure}
\footnotetext{\url{https://github.com/thunlp/OpenNRE}}

Table~\ref{tab:nyt_auc_precision_at_n} shows the results of our model on the held-out dataset. DISTRE with selective attention achieves a new state-of-the-art AUC value of $0.422$. 
The precision-recall curve in Figure~\ref{fig:nyt_pr_curve} shows that it outperforms RESIDE and PCNN+ATT at higher recall levels, while precision is lower for top predicted relation instances. The results of the PCNN+ATT model indicate that its performance is only better in the very beginning of the curve, but its precision drops early and only achieves an AUC value of $0.341$. Similar, RESIDE performs better in the beginning but drops in precision after a recall-level of approximately 0.25. This suggests that our method yields a more balanced overall performance, which we believe is important in many real-world applications.

Table~\ref{tab:nyt_auc_precision_at_n} also shows detailed precision values measured at different points along the P-R curve. We again can observe that while DISTRE has lower precision for the top 500 predicted relation instances, it shows a state-of-the-art precision of 60.2\% for the top 1000 and continues to perform higher for the remaining, much larger part of the predictions.
\begin{table*}[t!]
    \begin{center}
        \begin{tabular}{l c@{\hskip .3in} c c c c c c c}
            \toprule
            System & AUC & P@100 & P@200 & P@300 & P@500 & P@1000 & P@2000 \\
            \midrule
            Mintz$^\mathrm{\dagger}$ & 0.107 & 52.3 & 50.2 & 45.0 & 39.7 & 33.6 & 23.4 \\
            PCNN+ATT$^\mathrm{\ddagger}$ & 0.341 & 73.0 & 68.0 & 67.3 & 63.6 & 53.3 & 40.0 \\
            RESIDE$^\mathrm{\dagger}$ & 0.415 & \textbf{81.8} & \textbf{75.4} & \textbf{74.3} & \textbf{69.7} & 59.3 & 45.0 \\
            DISTRE & \textbf{0.422} & 68.0 & 67.0 & 65.3 & 65.0 & \textbf{60.2} & \textbf{47.9} \\
            \bottomrule
        \end{tabular}
        \caption{Precision evaluated automatically for the top rated relation instances. $\mathrm{\dagger}$ marks results reported in the original paper. $\mathrm{\ddagger}$ marks our results using the OpenNRE implementation.}
        \label{tab:nyt_auc_precision_at_n}
    \end{center}
\end{table*}

\subsection{Manual Evaluation and Analysis}
\label{subsec:manual_eval}
Since automated evaluation on a distantly supervised, held-out dataset does not reflect the actual performance of the models given false positive labels and incomplete knowledge base information, we also evaluate all models manually. This also allows us to gain a better understanding of the difference of the models in terms of their predictions. To this end, three human annotators manually rated the top 300 predicted relation instances for each model. Annotators were asked to label a predicted relation as correct only if it expressed a true fact at some point in time (e.g., for a \textit{/business/person/company} relationship, a person may have worked for a company in the past, but not currently), and if at least one sentence clearly expressed this relation, either via a syntactic pattern or via an indicator phrase. 

Table~\ref{tab:manual_eval} shows the \textit{P@100}, \textit{P@200}, \textit{P@300} and average precision scores, averaged over all annotators. PCNN+ATT has the highest average precision at $94.3\%$, $3\%$ higher than the $91.2\%$ of RESIDE and $5\%$ higher than our model. However, we see that this is mainly due to PCNN+ATT's very high \textit{P@100} and \textit{P@200} scores. For {P@300}, all models have very similar precision scores. PCNN+ATT's scores decrease considerably, reflecting the overall trend of its PR curve, whereas RESIDE's and DISTRE's manual precision scores remain at approximately the same level. Our model's precision scores for the top rated predictions are around 2\% lower than those of RESIDE, confirming the results of the held-out evaluation. Manual inspection of DISTRE's output shows that most errors among the top predictions arise from wrongly labeled \textit{/location/country/capital} instances, which the other models do not predict among the top 300 relations.

\begin{table*}[t!]
\centering
\begin{tabular}{lcccc}
\toprule
System & P@100 & P@200 & P@300 & Avg Prec \\ \midrule
PCNN+ATT & 97.3 & 94.7 & 90.8 & 94.3 \\
RESIDE & 91.3 & 91.2 & 91.0 & 91.2 \\
DISTRE & 88.0 & 89.8 & 89.2 & 89.0 \\ \bottomrule
\end{tabular}
\caption{Precision evaluated manually for the top 300 relation instances, averaged across 3 human annotators.}
\label{tab:manual_eval}
\end{table*}

Table \ref{tab:top_rel_distrib} shows the distribution over relation types for the top 300 predictions of the different models. 
We see that DISTRE's top predictions encompass 10 distinct relation types, more than the other two models, with \textit{/location/location/contains} and \textit{/people/person/nationality} contributing 67\% of the predictions. Compared to PCNN+ATT and RESIDE, DISTRE predicts additional relation types, such as e.g.\ \textit{/people/person/place\_lived} (e.g., "Sen. PER, Republican/Democrat of LOC") and \textit{/location/neighborhood/neighborhood\_of} (e.g., "the LOC neighborhood/area of LOC"), with high confidence. 
 
RESIDE's top 300 predictions cover a smaller range of 7 distinct relation types, but also focus on \textit{/location/location/contains} and \textit{/people/person/nationality} (82\% of the model's predictions). RESIDE's top predictions include e.g.\ the additional relation types \textit{/business/company/founders} (e.g., "PER, the founder of ORG") and \textit{/people/person/children} (e.g., "PER, the daughter/son of PER"). 
 
PCNN+ATT's high-confidence predictions are strongly biased towards a very small set of only four relation types. Of these, \textit{/location/location/contains} and \textit{/people/person/nationality} together make up 91\% of the top 300 predictions. Manual inspection shows that for these relations, the PCNN+ATT model picks up on entity type signals and basic syntactic patterns, such as "LOC, LOC" (e.g., "Berlin, Germany") and "LOC in LOC" ("Green Mountain College in Vermont") for \textit{/location/location/contains}, and "PER of LOC" ("Stephen Harper of Canada") for \textit{/people/person/nationality}. This suggests that the PCNN model ranks short and simple patterns higher than more complex patterns where the distance between the arguments is larger. The two other models, RESIDE and DISTRE, also identify and utilize these syntactic patterns.
 
\begin{table}[t!]
\centering
\begin{tabular}{lrrr}
\toprule
relation         & DIS &  RES &  PCNN           \\
\midrule
location/contains  &       168 &   182 &     214 \\
person/nationality &        32 &    65 &      59 \\
person/company     &        31 &    26 &      19 \\
person/place\_lived &        22 &     -- &       -- \\
country/capital    &        17 &     -- &       -- \\
admin\_div/country  &        13 &    12 &       6 \\
neighborhood/nbhd\_of       &        10 &     3 &       2 \\
location/team      &         3 &     -- &       -- \\
company/founders   &         2 &     6 &       -- \\
team/location      &         2 &     -- &       -- \\
person/children    &         -- &     6 &       -- \\
\bottomrule
\end{tabular}
\caption{Distribution over the top 300 predicted relations for each method. DISTRE achieves performance comparable to RESIDE, while predicting a more diverse set of relations with high confidence. PCNN+ATT shows a strong focus on two relations: \textit{/location/location/contains} and \textit{/people/person/nationality}.}
\label{tab:top_rel_distrib}
\end{table}
 
Table~\ref{tab:example_predictions} lists some of the more challenging sentence-level predictions that our system correctly classified.

\begin{table*}[t!]
\centering
\begin{tabular}{p{11cm}@{\hskip .5cm} l}
\toprule
Sentence & Relation \\
\midrule
Mr. Snow asked, referring to Ayatollah \textbf{Ali Khamenei}, \textbf{Iran}'s supreme leader, and Mahmoud Ahmadinejad, \textbf{Iran}'s president. & /people/person/nationality \\[20pt]
In \textbf{Oklahoma}, the Democratic governor, \textbf{Brad Henry}, vetoed legislation Wednesday that would ban state facilities and workers from performing abortions except to save the life of the pregnant woman. & /people/person/place\_lived \\[35pt]
\textbf{Jakarta} also boasts of having one of the oldest golf courses in \textbf{Asia}, Rawamangun , also known as the Jakarta Golf Club. &  /location/location/contains \\[20pt]
Cities like New York grow in their unbuilding: demolition tends to precede development, most urgently and particularly in \textbf{Lower Manhattan}, where \textbf{New York City} began. & /location/location/contains \\
\bottomrule
\end{tabular}
\caption{Examples of challenging relation mentions. These examples benefit from the ability to capture more complex features. Relation arguments are marked in bold.}
\label{tab:example_predictions}
\end{table*}

\section{Related Work}
\label{related_work}
\noindent  \textbf{Relation Extraction} \quad Initial work in RE uses statistical classifiers or kernel based methods in combination with discrete syntactic features, such as part-of-speech and named entities tags, morphological features, and WordNet hypernyms~\cite{mintz_distantsf_2009, hendrickx_semeval2010t8_2010}. These methods have been superseded by sequence based methods, including recurrent~\cite{socher_semanticct_2012, zhang_relationcv_2015} and convolutional neural networks~\cite{zeng_relationcv_2014, zeng_distant_2015}. Consequently, discrete features have been replaced by distributed representations of words and syntactic features~\cite{turian_wordra_2010, pennington_glove_2014}. \citet{xu_semanticrc_2015, xu_classifyingrv_2015} integrated shortest dependency path (SDP) information into a LSTM-based relation classification model. Considering the SDP is useful for relation classification, because it focuses on the action and agents in a sentence~\cite{bunescu_shortest_2005,socher_groundedcs_2014}. \citet{zhang_graphco_2018} established a new state-of-the-art for relation extraction on the TACRED dataset by applying a combination of pruning and graph convolutions to the dependency tree.
Recently, \citet{verga_transformer_2018} extended the Transformer architecture by a custom architecture for supervised biomedical named entity and relation extraction. In comparison, we fine-tune pre-trained language representations and only require distantly supervised annotation labels.

\noindent \textbf{Distantly Supervised Relation Extraction} \quad Early distantly supervised approaches~\cite{mintz_distantsf_2009} use multi-instance learning~\cite{riedel_modeling_2010} and multi-instance multi-label learning~\cite{surdeanu_multi_instance_2012, hoffmann_knowledge_based_2011} to model the assumption that at least one sentence per relation instance correctly expresses the relation. With the increasing popularity of neural networks, PCNN~\cite{zeng_relationcv_2014} became the most widely used architecture, with extensions for multi-instance learning~\cite{zeng_distant_2015}, selective attention~\cite{lin_neuralre_2016, re_relation_att}, adversarial training~\cite{wu_adversarial_re, qin_gan_re}, noise models~\cite{re_noise_matrix}, and soft labeling~\cite{re-soft-labels, label-free-distsup}. 
Recent work showed graph convolutions~\cite{reside_re} and capsule networks~\cite{emnlp_re_capsules}, previously applied to the supervised setting~\cite{zhang_graphco_2018}, to be also applicable in a distantly supervised setting. In addition, linguistic and semantic background knowledge is helpful for the task, but the proposed systems typically rely on explicit features, such as dependency trees, named entity types, and relation aliases~\cite{reside_re, re_noise_mitig}, or task- and domain-specific pre-training~\cite{re_ner_transfer, re-see}, whereas our method only relies on features captured by a language model during unsupervised pre-training.

\noindent \textbf{Language Representations and Transfer Learning} \quad
Deep language representations have shown to be an effective form of unsupervised pre-training. \citet{peters_deepcw_2018} introduced embeddings from language models (ELMo), an approach to learn contextualized word representations by training a bidirectional LSTM to optimize a disjoint bidirectional language model objective. Their results show that replacing static pre-trained word vectors~\cite{mikolov_efficienteo_2013,pennington_glove_2014} with contextualized word representations significantly improves performance on various natural language processing tasks, such as semantic similarity, coreference resolution, and semantic role labeling. \citet{ruder_universallm_2018} found language representations learned by unsupervised language modeling to significantly improve text classification performance, to prevent overfitting, and to increase sample efficiency. \citet{radford_improvinglu_2018} demonstrated that general-domain pre-training and task-specific fine-tuning, which our model is based on, achieves state-of-the-art results on several question answering, text classification, textual entailment, and semantic similarity tasks. \citet{devlin_bert_2018} further extended language model pre-training by introducing a slot-filling objective to jointly train a bidirectional language model. Most recently \cite{radford_gpt2} found that considerably increasing the size of language models results in even better generalization to downstream tasks, while still underfitting large text corpora.

\section{Conclusion}
We proposed DISTRE, a Transformer which we extended with an attentive selection mechanism for the multi-instance learning scenario, common in distantly supervised relation extraction.
While DISTRE achieves a lower precision for the 300 top ranked predictions, we observe a state-of-the-art AUC and an overall more balanced performance, especially for higher recall values. Similarly, our approach predicts a larger set of distinct relation types with high confidence among the top predictions.
In contrast to RESIDE, which uses explicitly provided side information and linguistic features, our approach only utilizes features implicitly captured in pre-trained language representations.
This allows for an increased domain and language independence, and an additional error reduction because pre-processing can be omitted.

In future work, we want to further investigate the extent of syntactic structure captured in deep language language representations. Because of its generic architecture, DISTRE allows for integration of additional contextual information, e.g. background knowledge about entities and relations, which could also prove useful to further improve performance.

% uncomment for final version
\section*{Acknowledgments}
We would like to thank the anonymous reviewers for their comments. This research was partially supported by the German Federal Ministry of Education and Research through the projects DEEPLEE (01IW17001) and BBDC2 (01IS18025E), and by the German Federal Ministry of Transport and Digital Infrastructure through the project DAYSTREAM (19F2031A).

\bibliography{references}
\bibliographystyle{acl_natbib}

\end{document}